\documentclass[11pt]{article}

\usepackage[preprint]{acl}

\usepackage{times}
\usepackage{latexsym}
\usepackage{multirow}
\usepackage{booktabs}
\usepackage{amsmath}
\usepackage{amssymb}

\usepackage[T1]{fontenc}

\usepackage[utf8]{inputenc}

\usepackage{microtype}

\usepackage{inconsolata}

\usepackage{graphicx}

\title{Simple Vision–Language Math Reasoning via Rendered Text}

\author{Matvey Skripkin \\
  FusionBrain Lab \\
  \texttt{skripkin@fusionbrainlab.com} \\\And
  Elizaveta Goncharova \\
  FusionBrain Lab \\
  HSE University \\\And
  Andrey Kuznetsov \\
  FusionBrain Lab \\
  Innopolis University \\
  }

\begin{document}
\maketitle
\begin{abstract}
We present a lightweight yet effective pipeline for training vision–language models to solve math problems by rendering \LaTeX{}‑encoded equations into images and pairing them with structured chain‑of‑thought prompts. This simple text‑to‑vision augmentation enables compact multimodal architectures to achieve state‑of‑the‑art reasoning accuracy. Through systematic ablations, we find that rendering fidelity and prompt design are the primary drivers of performance. Despite its simplicity, our approach consistently matches or surpasses both open‑source and proprietary math‑focused vision–language solvers on widely used benchmarks, while preserving broad general‑domain competence — showing gains on tasks such as MMMU, ChartQA, and DocVQA of up to 20\%. 
\end{abstract}

\section{Introduction}
\label{sec:intro}

Mathematical reasoning remains a core challenge for modern multimodal models. Existing vision–language architectures often struggle with (1) accurately parsing complex visual formulas and diagrams and (2) generating coherent chain-of-thought (CoT) explanations that mirror human problem-solving steps. In purely text-based settings, CoT supervision has been shown to significantly boost performance on a variety of math benchmarks~\cite{deepseekai2025,luo2025wizardmath}. However, publicly available multimodal datasets that provide both high-quality rendered math problems and step-by-step CoT annotations are scarce.

To address this gap, we propose a simple yet effective pipeline for converting large text-only math corpora into multimodal CoT training sets. Our method (i) samples problems from an existing benchmark, (ii) renders each problem --- including inline and display \LaTeX{} equations --- as an image, and (iii) pairs the image with a structured CoT prompt that first elicits intermediate reasoning steps and then the final answer. We perform ablation studies on compact vision–language models to isolate the contributions of rendering resolution and architectural details. Finally, we show that models trained on our generated dataset substantially outperform most math-focused counterparts and match the performance of state-of-the-art models of similar size trained on prior multimodal collections. This holds across standard vision-enhanced math benchmarks. Moreover, our model retains strong general capabilities, achieving a high \textsc{MMMU} score of 47.6 without explicit reasoning prompts --- whereas other models often overfit to chain-of-thought style and fail on basic multiple-choice questions (see Section~\ref{sec:experiments}).

Our contributions are threefold:
\begin{enumerate}
  \item A dataset construction pipeline for multimodal CoT supervision over text-only math problems.
  \item Detailed ablation studies that identify the most impactful design choices in rendering and prompt engineering.
\end{enumerate}

\section{Related Works}
\label{sec:related_works}

\paragraph{Mathematical Vision-Language Models.} Recent architectural innovations have produced specialized models demonstrating targeted approaches to mathematical reasoning. Math-LLaVA~\cite{shi2024mathllava} achieved a 19-point improvement over LLaVA-1.5 through systematic data synthesis, reaching 46.6\% on MathVista while maintaining lightweight architecture. G-LLaVA~\cite{gao2023gllava} focused on geometric problem solving, outperforming GPT-4V on MathVista's geometry subset (57.7\% vs 50.5\%) through specialized training on the Geo170K dataset. MathCoder-VL~\cite{wang-etal-2025-mathcoder} introduced code as an intermediate representation between mathematical images and formal descriptions, achieving 73.6\% accuracy on geometry problems through its FigCodifier component. These approaches demonstrate that focused architectural design can achieve competitive performance with significantly reduced computational requirements.

\paragraph{Multimodal Chain-of-Thought Reasoning.} The adaptation of chain-of-thought reasoning to multimodal contexts has emerged as a crucial development for mathematical VLMs.~\citet{zhang2024multimodalcot} established foundational work achieving state-of-the-art on ScienceQA with sub-1B parameter models through a two-stage framework separating rationale generation and answer inference. Multimodal Chain of Continuous Thought~\cite{pham2025multimodalcot} represents a significant efficiency innovation, reasoning directly in joint latent space rather than natural language tokens, achieving up to 8.23\% accuracy gains while reducing computational overhead. These advances in structured reasoning provide crucial foundations for lightweight systems that must balance reasoning capability with computational constraints. In addition, reinforcement learning techniques are successfully applied to the vision-language models to boost their geometry and general mathematical skills~\citep{huang2025visionr1,su2025openthink,chen2025reason}.

\paragraph{Mathematical Notation Processing.} Mathematical notation understanding has seen significant advances relevant to lightweight rendering applications. PosFormer~\cite{guan2024posformer} introduced position forest transformers for handwritten mathematical expression recognition, achieving substantial gains on CROHME benchmarks through joint optimization of expression and position recognition. The Graph-Encoder-Transformer-Decoder approach~\cite{graph} demonstrated competitive results through explicit modeling of spatial relationships between symbols, directly converting to LaTeX format. MathWriting~\cite{gervais2025math} provides comprehensive LaTeX ground truth for both online and offline recognition. These advances in efficient mathematical notation processing are essential for lightweight systems that must parse and understand mathematical content without heavy computational requirements.




\section{MathVLM Tuning}

\subsection{Vision-Language Architecture}
\label{math-vlm}

To handle both general visual content and mathematical notation, we propose a multi-encoder fusion backbone incorporating several vision encoders into the large language model via linear projectors. We experimented with three vision encoders --- a general-purpose Vision Transformer (InternViT~\cite{chen2024internvlscalingvisionfoundation}), a \LaTeX-specialized transformer (Texify\footnote{https://github.com/VikParuchuri/texify}), and a high-resolution convolutional network (ConvNeXt~\cite{liu2022convnet2020s}) --- and fed their fused output into the lightweight LLM Qwen2.5~\cite{qwen2025qwen25technicalreport}.

\subsection{Visual Encoders and Fusion Strategies}

We employ three complementary visual encoders, each operating on differently sized inputs to capture distinct aspects of the data:
\begin{itemize}
  \item \textbf{InternViT}: A Vision Transformer pretrained on large-scale natural images, processing $448\times448$-pixel inputs to extract global image semantics.
  \item \textbf{Texify}: A transformer architecture trained on rendered \LaTeX{} snippets, optimized for accurate symbol layout and equation structure extraction from $420\times420$-pixel inputs.
  \item \textbf{ConvNeXt}: A pure convolutional network inspired by ViT design principles, demonstrating competitive performance on $1024\times1024$-pixel inputs and capturing fine-grained visual details.
\end{itemize}
Each input image $I$ is resized independently for each encoder. In our experiments, we consider two encoder pairings: InternViT+Texify and InternViT+ConvNeXt. To integrate their outputs into the LLM backbone (Qwen2.5 1.5B or 7B), we propose two fusion strategies:

\paragraph{Sequence-level Fusion (InternViT + Texify).}  
We first obtain embeddings $e_I\in\mathbb{R}^{l_I \times d_I}$ and $e_T\in\mathbb{R}^{l_T \times d_T}$ from InternViT and Texify, respectively. Each embedding is projected into the LLM’s token space via dedicated adapter matrices:
\[
z_I = e_I\,W_I,\quad
z_T = e_T\,W_T,
\]
where $W_I\in\mathbb{R}^{d_I\times d_{\mathrm{LLM}}}$ and $W_T\in\mathbb{R}^{d_T\times d_{\mathrm{LLM}}}$. We concatenate them along the sequence axis:
\[
z = [\,z_I; z_T\,]\;\in\;\mathbb{R}^{(\ell_I + \ell_T)\times d_{\mathrm{LLM}}}.
\]
This extended token sequence is prepended to the LLM's input embeddings, enabling independent attention over both natural-image features and LaTeX-specific structural cues.

This extended sequence of image embeddings is then prepended to the textual tokens embeddings, thus allowing the LLM to process signal from several encoders simoulteneously.

\paragraph{Feature-level Fusion (InternViT + ConvNeXt).}  
For the InternViT+ConvNeXt pairing, we first merge their raw embeddings $e_I\in\mathbb{R}^{l_I \times d_I}$ and $e_C\in\mathbb{R}^{l_C \times d_C}$ by concatenation:
\[
e_{IC} = [\,e_I;\,e_C\,]\;\in\;\mathbb{R}^{l \times (d_I + d_C)}.
\]
This fused vector is then projected through a single adapter matrix,
\[
z = e_{IC}W_F,,\quad
W_F\in\mathbb{R}^{(d_I + d_C)\times d_{\mathrm{LLM}}},
\]
producing a sequence of $l$ fusion tokens (each of dimension $d_{\mathrm{LLM}}$). We prepend these $l$ tokens to the LLM's input embeddings, maintaining a constant total sequence length while enabling joint attention over both global and fine-grained convolutional features.

\subsection{Datasets}

One key challenge in unifying mathematical reasoning with general visual understanding is assembling effective pretraining and fine-tuning data. We address this by curating specialized corpora for adapter alignment and supervised fine-tuning.

\paragraph{Adapter pre-training}  
We train two separate adapter modules when using InternViT+Texify — one to project InternViT's features and one for Texify's — whereas for InternViT+ConvNeXt we use a single shared adapter to map the concatenated features into the LLM's embedding space. These are due to the similarity of the embedding dimensions provided by InternViT and ConvNeXt encoders. Pre-training uses three multimodal corpora with strong captioning and OCR signals: TextCaps~\cite{sidorov2020textcapsdatasetimagecaptioning}, ShareGPT4V~\cite{chen2023sharegpt4vimprovinglargemultimodal}, and LLaVAR~\cite{zhang2024llavarenhancedvisualinstruction}. 

\begin{figure}
    \centering
    \includegraphics[width=0.8\linewidth]{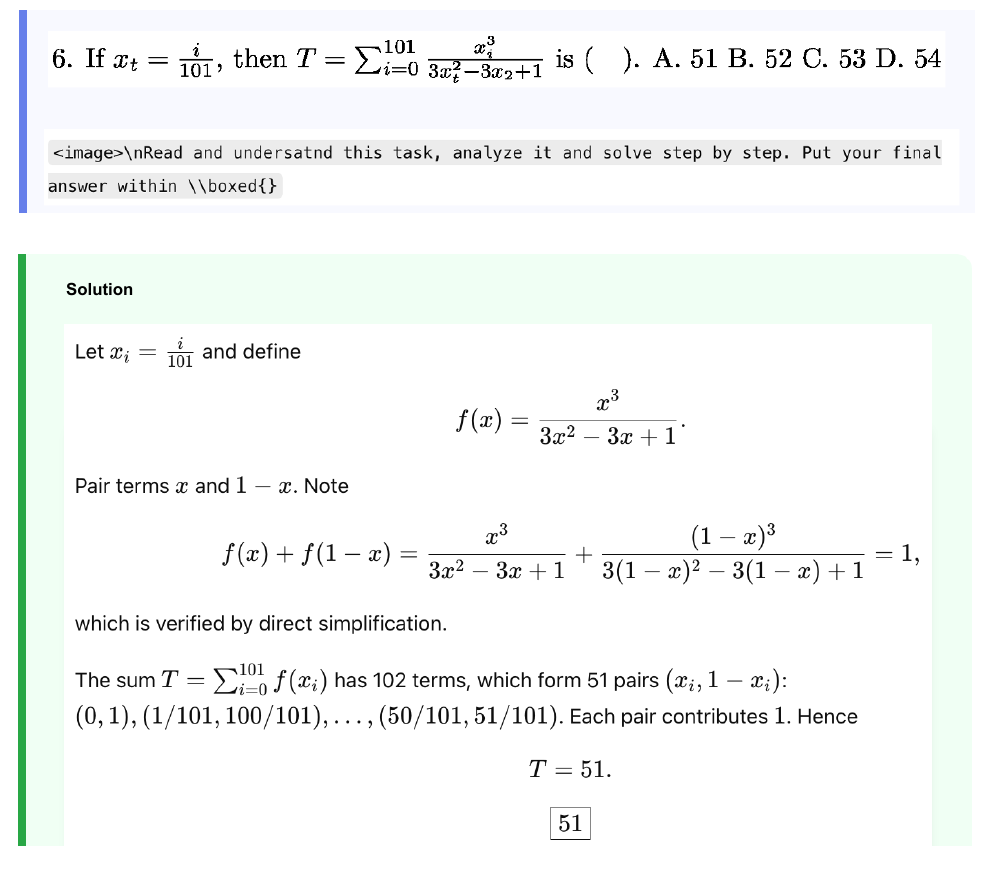}
    \caption{Example of the rendered math problem with the solution. First part is provided as an image for the VLM, while the solution is in \LaTeX}
    \label{fig:rend-ex}
\end{figure}

\paragraph{Supervised fine-tuning}  
We build the \emph{Open‑R1‑Rendered} dataset by rasterizing 93.7k Open‑R1 text‑only math problems~\cite{openr1} (text and \LaTeX source) into images at multiple resolutions and pairing each image with its original prompt and chain‑of‑thought (CoT) solution. Figure~\ref{fig:rend-ex} provides an example of the source text and the rendered image used for training. By exposing the model to both the visual rendering and the CoT supervision, the dataset improves visual CoT reasoning while simultaneously developing OCR‑like skills, which translate into better performance on image‑based reading and reasoning.

To assess the impact of explicit \LaTeX parsing, we supply two variants (image + \LaTeX formulation + solution vs. image + solution) and compare their performance in our ablation study (Section~\ref{sec:abl}).

We then augment Open-R1-rendered dataset with six domain-specific datasets:
\begin{itemize}
  \item \textbf{Cambrian737K}~\cite{tong2024cambrian1fullyopenvisioncentric} includes 665K instruction-following samples from LLaVA-1.5, enriched with OCR outputs and chart-related tasks. Its diverse captions and layout variations help the model generalize across text-and-vision instructions.
  \item \textbf{MultiMath-300K}~\cite{peng2024multimathbridgingvisualmathematical} contains 300K vision-language examples at K-12 levels-arithmetic, algebra, geometry, and functional analysis. Chain-of-thought annotations teach the model to break problems into intermediate steps.
  \item \textbf{GeoQA-170K}~\cite{chen2022geoqageometricquestionanswering} focuses on geometric diagram problems, providing 170K questions with step-by-step solutions. This dataset sharpens spatial reasoning and diagram interpretation.
  \item \textbf{MathV-360K}~\cite{shi2024mathllavabootstrappingmathematicalreasoning} offers 360K competition-sourced math problems across 16 disciplines and five difficulty tiers. Its varied problem styles build robustness to different reasoning complexities.
  \item \textbf{SciGraphQA-295K}~\cite{li2023scigraphqalargescalesyntheticmultiturn} is a synthetic, multi-turn QA set over academic graph images. More complex than ChartQA, it challenges the model to track context across turns and interpret real-world graph layouts.
  \item \textbf{MMC Instruction-400K}~\cite{liu2024mmcadvancingmultimodalchart} contains 400K chart interpretation tasks with diverse chart types and vision-centric prompts. It reinforces fine-grained chart reading and multimodal instruction following.
  \item \textbf{\LaTeX{}-300K} contains formulas alongside images to strengthen symbolic reasoning.
\end{itemize}

\begin{figure}
    \centering
    \includegraphics[width=0.8\linewidth]{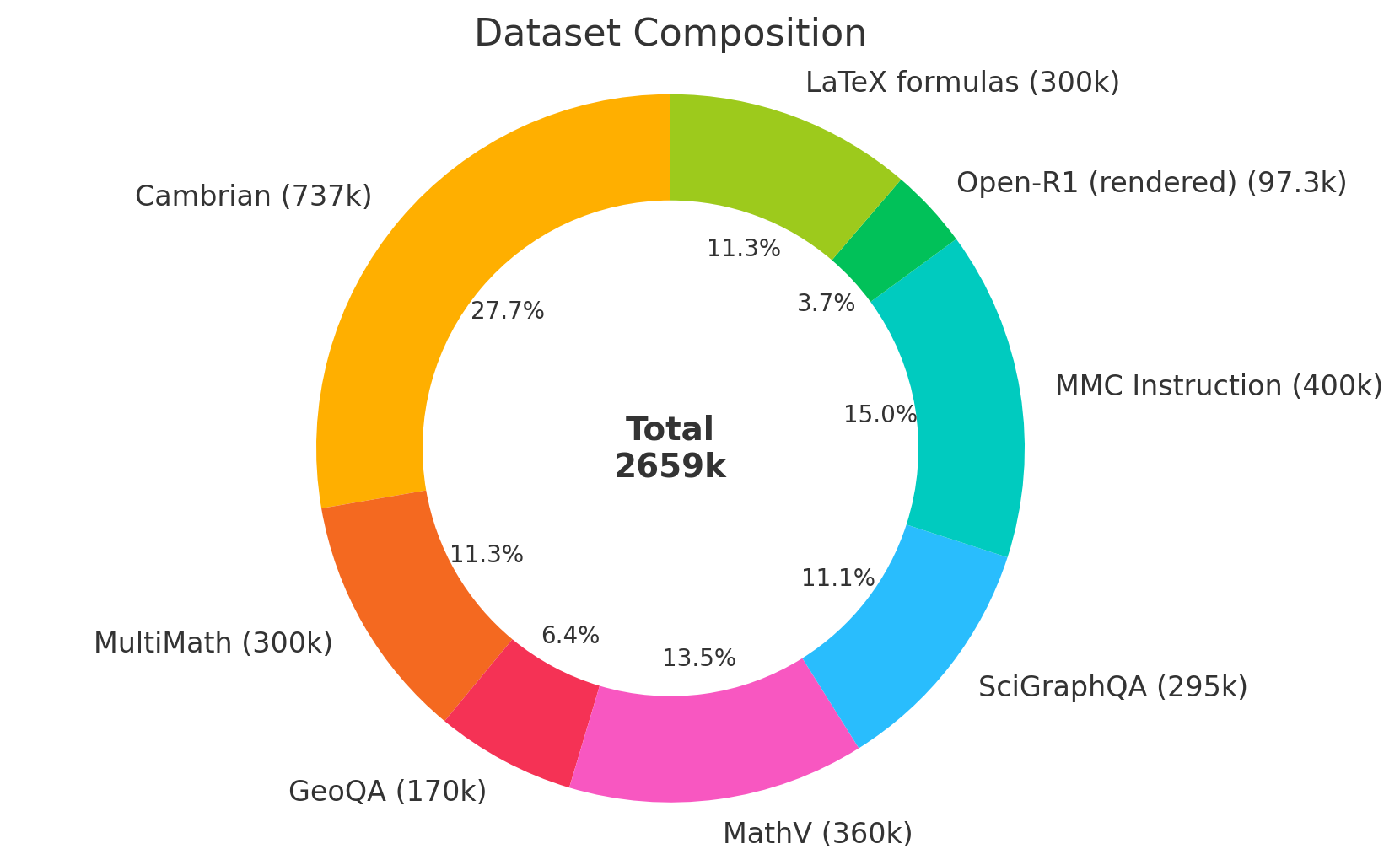}
    \caption{SFT training data distribution}
    \label{fig:enter-label}
\end{figure}

\subsection{Training Procedure}

We adopted a two‑stage training strategy. In the adapter‑only stage, we froze the base LLM and trained only the adapters on the task data. In the subsequent supervised fine‑tuning (SFT) stage, we unfroze the LLM and jointly fine‑tuned the LLM and adapters. In our final experiment, we additionally unfroze the encoders and trained them jointly with the LLM.

With the encoders unfrozen, we train for two epochs on the SFT dataset; its distribution is shown in Figure~\ref{fig:enter-label}. Training uses eight A100 80GB GPUs and a cosine learning‑rate schedule with base learning rates of 1e‑3 and 2e‑5.

\section{Experiments}
\label{sec:experiments}

In this section, we present the experimental evaluation of the proposed architecture on various mathematical tasks, demonstrating the models' performance on tasks related to the architecture.


\begin{table*}[ht]
\centering
\caption{Evaluation of the various models' performance on mathematical-based tasks.}
\label{tab:vlm-general}
\resizebox{\textwidth}{!}{%
\begin{tabular}{lcccccccc}
\toprule
& \textbf{Size} & \textbf{Avg} & \textbf{MathVerse} & \textbf{MathVision} & \textbf{MathVista} & \textbf{WE-MATH} & \textbf{DYNAMATH} & \textbf{LogicVista} \\
& & & testmini & full set & gps & & & \\
\midrule
\multicolumn{9}{l}{\textbf{Closed-Source MLLMs}} \\
\midrule
GPT-4o          & -   & 54.6 & 50.2 & 30.4 & 64.7 & 62.8 & 64.9 & — \\
Gemini-1.5-pro & -   & 52.7 & 35.3 & 19.2 & 81.7 & 66.9 & 60.5 & — \\
\midrule
\multicolumn{9}{l}{\textbf{Open-Source General MLLMs}} \\
\midrule
InternVL-Chat-V1.5            & 26B & 33.6 & 26.1 & 15.4 & 56.9 & 32.7 & 36.7 & — \\
Llama-3.2-11B-Vision-Instruct & 11B & 26.2 & 28.9 & 16.9 & 40.9 & 12.0 & 32.2 & — \\
Qwen2-VL                       & 8B  & 37.8 & 33.6 & 19.2 & 51.0 & 43.0 & 42.1 & — \\
InternVL2-8B                   & 8B  & 39.5 & 37.0 & 18.4 & 57.7 & 44.9 & 39.7 & — \\
InternVL2-8B-MPO               & 8B  & 42.9 & 38.2 & 22.3 & 69.2 & 44.4 & 40.5 & — \\
InternVL2.5-8B                 & 8B  & 41.9 & 39.5 & 19.7 & 64.9 & 44.7 & 40.5 & 36.0 \\
LLaVA-OneVision                & 8B  & 40.2 & 28.9 & 18.3 & 71.6 & 44.9 & 37.5 & 33.3 \\
Points-Qwen2.5-Instruct        & 8B  & 47.0 & 41.1 & 23.9 & 76.0 & 51.0 & 42.8 & — \\
Gemma3-12B                     & 12B & 46.1 & 40.1 & \textbf{29.1} & 63.6 & 51.7 & \textbf{45.8} & — \\
\midrule
\multicolumn{9}{l}{\textbf{Open-Source Reasoning MLLMs}} \\
\midrule
Math-LLaVA          & 13B & 32.6 & 22.9 & 15.7 & 57.7 & 31.3 & 35.5 & — \\
MathPUMA-Qwen2-7B   & 8B  & 34.8 & 33.6 & 14.0 & 48.1 & 41.0 & 37.3 & — \\
MultiMath           & 7B  & 38.2 & 27.7 & 16.3 & 66.8 & 42.2 & 37.9 & — \\
MAVIS               & 7B  & 39.7 & 35.2 & 18.5 & 64.1 & 44.3 & 36.2 & — \\
InfMM-Math          & 7B  & 44.6 & 40.5 & 18.8 & 77.3 & 48.3 & 38.2 & — \\
AtomThink-EMOVA     & 8B  & 46.7 & 42.5 & 24.9 & 75.9 & 49.3 & 40.9 & — \\
MathGLM-Vision      & 9B  & 43.0 & 44.2 & 19.2 & 64.2 & 45.2 & 42.2 & — \\
Llama-v-o1          & 11B & 36.5 & 33.9 & 17.9 & 53.3 & 42.6 & 34.7 & — \\
OpenVLThinker       & 7B  &  —   & \textbf{47.9} & 25.3 & 76.4 &  —   &  —   & — \\
R1-Onevision        & 7B  &  —   & 47.4 & 26.9 & 72.4 & 51.4 &  —   & — \\
URSA-8B             & 8B  & \textbf{50.9} (47.7 w. LogicVista)& 45.7 & 28.7 & \textbf{81.7} & 53.6 & 44.7 & 31.8 \\
\midrule
Math-VLM            & 8B  & 48.5 (46.8 w. LogicVista)& 41.4 & 23.9 & 76.0 & \textbf{60.6} & 40.6 & \textbf{38.3} \\
\bottomrule
\end{tabular}
}
\end{table*}


\begin{table*}[ht]
\centering
\footnotesize
\caption{Evaluation of the math-oriented models on general-domain benchmarks. Avg is the unweighted mean of ChartQA, DocVQA, InfoVQA, and MMMU.}
\label{tab:mllm_performance_gen}
\begin{tabular}{@{}lcccccc@{}}
\toprule
\textbf{Model} & \textbf{Size} & \textbf{Avg} & \textbf{ChartQA} & \textbf{DocVQA} & \textbf{InfoVQA} & \textbf{MMMU} \\
\midrule
Math-LLaVA          & 13B & 37.5 & 44.5 & 32.6 & 36.8 & 36.1 \\
MathPUMA-Qwen2-7B   & 8B  & 38.2 & 43.2 & 39.4 & 35.1 & 35.0 \\
MultiMath           & 7B  & 31.1 & 41.4 & 27.9 & 36.2 & 18.9 \\
URSA-8B             & 8B  & 31.6 & 43.7 & 29.9 & 20.0 & 32.8 \\
\midrule
Math-VLM            & 8B  & \textbf{59.9} & \textbf{73.0} & \textbf{76.4} & \textbf{42.7} & \textbf{47.6} \\

\bottomrule
\end{tabular}
\end{table*}

\begin{table*}[ht]
\centering
\small
\setlength{\tabcolsep}{4pt}
\caption{Model performance on the detailed WeMath benchmark. \textbf{Best scores within the \emph{Open-source Math MLLMs} block are bolded.}}
\label{tab:wemath}
\resizebox{\textwidth}{!}{%
\begin{tabular}{@{} l c r r r r r r r r r r r r r r r @{}}
\toprule
\multirow{2}{*}{\textbf{Model}} & \multirow{2}{*}{\textbf{\#Params}} & \multicolumn{3}{c}{\textbf{S}} &
  \multicolumn{2}{c}{\textbf{Mem}} &
  \multicolumn{2}{c}{\textbf{PF}} &
  \multicolumn{2}{c}{\textbf{SF}} &
  \multicolumn{2}{c}{\textbf{TMF}} &
  \multicolumn{4}{c}{\textbf{PD}} \\
\cmidrule(lr){3-5} \cmidrule(lr){6-7} \cmidrule(lr){8-9} \cmidrule(lr){10-11} \cmidrule(lr){12-13} \cmidrule(lr){14-17}
 & & \textbf{S1} & \textbf{S2} & \textbf{S3} & \textbf{UCU} & \textbf{AL} & \textbf{CPF} & \textbf{UPF} & \textbf{CSF} & \textbf{USF} & \textbf{BTF} & \textbf{CCF} & \textbf{Dir} & \textbf{Pos} & \textbf{RoM} & \textbf{CCP} \\
\midrule
\multicolumn{17}{l}{\textit{Closed-source MLLMs}}\\
\midrule
GPT-4o & - & 72.8 & 58.1 & 43.6 & 86.6 & 39.1 & 77.4 & 71.6 & 84.5 & 62.3 & 58.7 & 69.4 & 93.1 & 72.7 & 47.5 & 73.3 \\
GPT-4V & - & 65.5 & 49.2 & 38.2 & 82.5 & 38.4 & 70.7 & 60.2 & 76.6 & 56.3 & 57.8 & 67.7 & 79.3 & 57.5 & 47.8 & 63.3 \\
Gemini-1.5-Pro & - & 56.1 & 51.4 & 33.9 & 51.0 & 31.2 & 61.8 & 45.0 & 70.0 & 57.5 & 39.2 & 62.7 & 68.8 & 54.1 & 40.7 & 60.0 \\
Qwen-VL-Max & - & 40.8 & 30.3 & 20.6 & 19.4 & 25.3 & 39.8 & 41.4 & 43.6 & 48.0 & 43.8 & 43.4 & 41.4 & 35.1 & 40.7 & 26.7 \\
\midrule
\multicolumn{17}{l}{\textit{Open-source General MLLMs}}\\
\midrule
LLaVA-1.6 (7B)  & 7B  & 23.0 & 20.8 & 15.8 & 18.5 & 20.5 & 16.9 & 29.6 & 15.6 & 18.6 & 42.7 & 24.1 & 17.6 & 43.3 & 28.9 & 26.7 \\
LLaVA-1.6 (13B) & 13B & 29.4 & 25.3 & 32.7 & 21.7 & 23.2 & 23.4 & 34.7 & 25.3 & 26.4 & 37.5 & 41.7 & 26.9 & 28.9 & 37.1 & 30.0 \\
GLM-4V-9B       & 9B  & 47.3 & 37.2 & 38.2 & 53.4 & 37.0 & 51.3 & 46.5 & 50.6 & 38.2 & 44.1 & 45.2 & 41.0 & 49.3 & 36.8 & 53.3 \\
InternVL2.5-8B  & 9B  & 58.7 & 43.1 & 38.8 & 48.7 & 35.8 & 65.5 & 54.5 & 62.3 & 61.5 & 47.8 & 60.3 & 79.0 & 64.0 & 51.1 & 63.3 \\
Qwen2-VL        & 8B  & 59.1 & 43.6 & 26.7 & 62.7 & 37.2 & 62.6 & 60.8 & 65.7 & 49.2 & 52.5 & 49.2 & 48.1 & \textbf{68.2} & 55.0 & 56.7 \\
Gemma3-12B      & 12B & 64.3 & 47.2 & 42.1 & \textbf{83.1} & 33.9 & 70.2 & 58.2 & \textbf{77.5} & 61.1 & 50.1 & \textbf{63.7} & 82.6 & 58.4 & 36.8 & 60.0 \\
\midrule
\multicolumn{17}{l}{\textit{Open-source Math MLLMs}}\\
\midrule
G-LLaVA & 7B & 32.4 & 30.6 & 32.7 & 33.3 & 29.1 & 32.0 & 37.9 & 19.6 & 33.5 & 37.1 & 32.8 & 31.2 & 33.2 & 25.6 & 40.0 \\
Math-LLaVA & 13B & 38.7 & 34.2 & 34.6 & 30.3 & 17.9 & 39.2 & 40.4 & 37.1 & 37.7 & 53.0 & 51.3 & 30.8 & 30.8 & 40.9 & 46.7 \\
Math-PUMA-Qwen2-7B  & 8B & 53.3 & 39.4 & 36.4 & 63.5 & 42.5 & 60.2 & 45.9 & 66.2 & 48.6 & 42.3 & 53.5 & 31.2 & 37.7 & 40.4 & 46.7 \\
URSA-8B & 8B & 63.1 & 56.4 & 41.8 & 59.1 & 32.5 & 72.3 & 60.3 & 70.9 & \textbf{66.0} & 51.4 & 59.8 & 58.3 & 39.5 & \textbf{58.8} & 53.3 \\
\midrule
MathVLM & 8B & \textbf{69.1} & \textbf{60.6} & \textbf{50.3} & 71.4 & \textbf{44.4} & \textbf{75.9} & \textbf{67.1} & 73.4 & 61.7 & \textbf{68.8} & 62.6 & \textbf{86.2} & 59.4 & 55.5 & \textbf{80.0} \\
\bottomrule
\end{tabular}%
}
\end{table*}

\subsection{Benchmarks}

For the evaluation, we chose MathVista~\cite{lu2024mathvista}, MathVerse~\cite{zhang2024mathverse}, DYNAMATH~\cite{zou2025dynamathdynamicvisualbenchmark},
WE-MATH~\cite{qiao2024wemathdoeslargemultimodal}, GeoQA~\cite{chen2022geoqageometricquestionanswering}, MathVision\cite{wang2024measuring}, and LogicVista~\cite{xiao2024logicvista}, as long as standard vision-language models benchmarks, including DocVQA~\cite{mathew2021docvqa}, ChartQA~\cite{masry2022chartqa}, InfoVQA~\cite{mathew2021infographicvqa}, and MMMU~\cite{yue2024mmmu}. While the first group of the benchmarks serve as the task to check the model's reasoning abilities in mathematical-based tasks, we use the second one estimate model's abilities to solve general-domain tasks.

Because several math-tuned models produce step-by-step reasoning rather than a short final span on general-domain tasks (DocVQA, ChartQA, InfoVQA, MMMU), we post-process their outputs by extracting the final answer with an LLM (DeepSeek-V3~\cite{deepseekai2025deepseekv3technicalreport}) and evaluate that span. Without this extraction, exact-match style metrics would assign a score of zero to long, reasoned outputs due to evaluation specifics. The example of such generations is given in Appendix~\ref{sec:appendix}. All evaluations are conducted using the LMMs-Eval framework~\cite{zhang2024lmmsevalrealitycheckevaluation,lmms_eval2024}.

\subsection{Ablation Study}

To identify the optimal training pipeline for our math‐focused modules, we first perform the data- and architectural ablations.

\subsubsection{Dataset Ablations}
\label{sec:abl}

For the data ablations, we used small Qwen 2.5 1.5B model as LLM backbone and standard architecture with InternViT + Texify encoders. By evaluating different dataset configurations on this lightweight backbone, we select the best-performing design before scaling to larger LLMs. The dataset mixes are shown in Table~\ref{tab:ablation-small-vision-0-5}.

The ablation evaluation setup is given below:

\begin{enumerate}
  \item \textbf{Rendered:} Cambrian737K + VisualMath + Open-R1-rendered + Unimer LaTeX
  \item \textbf{Text-only + TexTeller:} Cambrian737K + VisualMath + Open-R1 + TexTeller
  \item \textbf{Rendered + Geo Mavis:} Rendered mix (as above) + Geo Mavis MultiMath pretraining
\end{enumerate}


\begin{table}[h!]
  \centering
  \small
  \caption{Datasets ablation results for small vision modules. The best-performing metrics are given in bold, and second best-performing are underlined.}
  \label{tab:ablation-small-vision-0-5}
  \resizebox{\columnwidth}{!}{%
    \begin{tabular}{@{}l r r r@{}}
      \toprule
      \textbf{Dataset Description}      & \textbf{DynaMath Avg} & \textbf{MathVision Acc} & \textbf{WeMath (Strict)} \\
      \midrule
      Rendered      & \textbf{25.55} & 20.39                     & 12.76 \\
      OpenR1 (text-only) \& TexTeller LaTeX      & 19.28                              & 21.38                     & 18.95                    \\
      Add math alignment data (pretrain with Geo Mavis MultiMath)          & \underline{24.79}                              & \textbf{21.71} & 18.10                    \\
      \bottomrule
    \end{tabular}%
  }
\end{table}

Table~\ref{tab:ablation-small-vision-0-5} shows that full rendering (Model0) maximizes DynaMath but underperforms on WeMath strict-match, while the text-only + TexTeller setup (Model1) trades DynaMath and MathVision for better WeMath. Adding Geo Mavis pretraining to rendered data (Model2) yields a balanced trade-off, and is used for all subsequent experiments.

\subsubsection{Architectural Ablations}

Our architectural ablations on Qwen2.5-7B (Table~\ref{tab:arch-ablations-7B-colored}) show that \textbf{InternViT+Texify} achieves the best DynaMath score (41.26), however, \textbf{InternViT+ConvNeXt} with frozen encoders underperforms. Thus, we choose the following architecture InternViT+ConvNeXt with for final experiments on best-performing dataset.

\begin{table}[h!]
  \centering
  \small
  \caption{Architectural ablations on the 7B-parameter model}
  \label{tab:arch-ablations-7B-colored}
  \resizebox{\columnwidth}{!}{%
    \begin{tabular}{@{}l r r r@{}}
      \toprule
      \textbf{Architecture Variant} 
        & \textbf{DynaMath Avg} 
        & \textbf{MathVision Acc} 
        & \textbf{WeMath (Strict)} \\
      \midrule
      InternViT + TexiFy             & \textbf{41.26} & 22.04                     & 28.10                     \\
      InternViT + ConvNext   & \underline{40.10}                              & \textbf{24.01} & \textbf{31.43}                     \\
      \bottomrule
    \end{tabular}%
  }
\end{table}

\subsubsection{Experimental Results}
\label{sec:experiments}

We present results for the final architecture in Table~\ref{tab:vlm-general}. Among open-source generalists, our \textsc{MathVLM} outperforms comparably sized models on mathematical reasoning benchmarks. Against math-focused systems, it rivals the strong URSA baseline while surpassing many prior math-oriented models; notably, it achieves a leading score on \textsc{WE-MATH} (60.6) Score (Loose).

We also compare math-tuned models on general-domain tasks where reasoning is not required. As shown in Table~\ref{tab:mllm_performance_gen}, \textsc{MathVLM} attains the highest average (59.9) across ChartQA, DocVQA, InfoVQA, and MMMU, with strong single-task results (e.g., 73.0 on ChartQA and 76.4 on DocVQA). These gains indicate better prompt adherence and robustness on short-form answer extraction, suggesting that our synthesized vision–language CoT data not only improves math reasoning but also yields more balanced multimodal behavior.

Table~\ref{tab:wemath} details performance on \textsc{WeMath}. Accuracy decreases from one- to three-step problems (S1$\rightarrow$S3: 69.1$\rightarrow$50.3), highlighting the challenge of multi-step visual reasoning. Among open-source math models, \textsc{MathVLM} achieves the highest overall mean and leads most step-based groups, while also topping several skill domains (e.g., TMF: 68.8/62.6 for BTF/CCF; PD: 86.2 on Dir and 80.0 on CCP). URSA remains particularly strong on solid-figure geometry, but \textsc{MathVLM} is the most balanced and highest-scoring model overall.

\begin{table}[t]
\centering
\footnotesize
\caption{Benchmark results (mean~$\pm$~std) for a single prompt over three runs.}
\label{tab:benchmark_single_prompt}
\resizebox{\linewidth}{!}{%
\begin{tabular}{@{}lcccc@{}}
\toprule
\textbf{Benchmark} & \textbf{MathVision} & \textbf{MathVista} & \textbf{WeMath} & \textbf{DynaMath} \\
\midrule
Version 3 & $23.77 \pm 0.11$ & $75.96 \pm 0.00$ & $60.67 \pm 0.08$ & $40.55 \pm 0.03$ \\
\bottomrule
\end{tabular}%
}
\end{table}

\subsubsection{Prompt Ablations}

We also present the ablation study of the dependency of model's performance on the benchmarks based on the input prompt. The list of the input prompts is given in Table~\ref{tab:prompt_types}, while the results with different prompts fo our VLM model is given in Table~\ref{tab:prompt_ablation}.

\begin{table*}[h!]
\centering
\footnotesize
\caption{Prompt composition strategies for multimodal inputs with image token and question.}
\label{tab:prompt_types}
\begin{tabular}{@{}lll@{}}
\toprule
\textbf{Prompt Variant} & \textbf{Structure} & \textbf{Description} \\
\midrule
Between        & [IMG] + suffix + question      & Suffix is inserted between the image token and the question. \\
Before         & suffix + [IMG] + question      & Suffix is placed before the image token. \\
After          & [IMG] + question + suffix      & Suffix is appended after the question. \\
\bottomrule
\end{tabular}
\end{table*}

\begin{table*}[h!]
\centering
\footnotesize
\caption{Different suffix versions}
\label{tab:prompt_types}
\begin{tabular}{@{}l p{12cm} @{}}
\toprule
\textbf{Suffix Variant} & \textbf{Description} \\
\midrule
Version 1        & You are given a math problem image, read and understand this task, analyze it, and provide step by step solution. \\
Version 2        & You are given an image containing a math problem. Read the image, identify the problem statement and all important data, then produce a clear step-by-step solution and the final answer. Use concise steps and label the final answer. \\
Version 3        & You are an expert math tutor. Your goal is to help a student understand the problem in the image. Carefully examine the task. Provide a detailed, step-by-step solution. Explain the logic behind each step in simple and clear language. Make sure your explanation helps to understand the topic, not just to get the answer. At the end, highlight the final answer. \\
\bottomrule
\end{tabular}
\end{table*}

\begin{table*}[h!]
\centering
\footnotesize
\caption{Effect of different prompt formulations on performance across math-related benchmarks.}
\label{tab:prompt_ablation}
\resizebox{\columnwidth}{!}{%
\begin{tabular}{@{}lccccc@{}}
\toprule
\textbf{Prompt Variant} & \textbf{Avg} & \textbf{MathVision} & \textbf{MathVista gps} & \textbf{WeMath} & \textbf{DynaMath} \\
\midrule
Without suffix          & 47.8 & 22.8 & 70.2 & 57.5 & 40.7 \\
Version 1 (before)      & 48.4 & 22.9 & 70.7 & 57.7 & \textbf{42.4} \\
Version 1 (after)       & 49.7 & \textbf{24.3} & 71.6 & \textbf{61.8} & 41.1 \\
Version 1 (between)     & 48.5 & 22.9 & 68.8 & 60.3 & 41.9 \\
Version 2 (after)       & 49.8 & 23.0 & 73.6 & 61.1 & 41.3 \\
Version 3 (after)       & \textbf{50.3} & 23.9 & \textbf{76.0} & 60.6 & 40.6 \\
\bottomrule
\end{tabular}
}
\end{table*}

\section{Discussion}
\label{sec:discussion}

The obtained results shows that adding high-quality rendered dataset with mathematical CoT reasoning can boost model's performance on the VLM reasoning benchmark. We show that on math-heavy problems, MathVLM outperforms most general-domain models, even with bigger LLM backbone, however, keeping their ability to solve general tasks in the same manner.

\paragraph{Trade-offs versus text-only CoT.}
On WeMath, performance degrades from one- to three-step problems (S1$\rightarrow$S3), consistent with prior findings that multi-step visual reasoning compounds small perception errors. Text-only supervision plus TexTeller improves strict-match on measurement tasks, but underperforms on visual composition tasks (charts, diagrams), suggesting that visual grounding is necessary for layout-heavy domains.

\paragraph{Fusion design choices.}
Although InternViT+Texify slightly improves DynaMath, we selected InternViT+ConvNeXt for the final model due to better strict-match on WeMath and stronger robustness on DocVQA/ChartQA. In practice, ConvNeXt contributes fine-grained sensitivity to details and line-detail sensitivity, while InternViT supplies global context. We hypothesize that Texify’s benefits overlap with rendering quality; when render fidelity is high, generic encoders suffice.

\paragraph{Stability of Model Evaluation}

Because most math-focused benchmarks are scored by an LLM judge, we also assess run-to-run stability. Table~\ref{tab:benchmark_single_prompt} reports the mean~$\pm$~standard deviation over three independent evaluations of a single prompt. Decoding is greedy, so variation primarily reflects the LLM judge rather than sampling. Across benchmarks, scores are highly stable, varying by at most $0.11$ percentage points.

\section{Conclusion}
\label{sec:conclusion}

We presented a simple pipeline that converts text‑only math problems into a vision–language dataset by rendering \LaTeX{} to images and pairing each image with structured chain‑of‑thought (CoT) supervision. Through targeted ablations, we showed that \emph{rendering fidelity} and \emph{CoT prompt design} are the dominant factors driving reasoning performance, with fusion strategy also affecting robustness across tasks. Models trained on the synthesized data achieve competitive results on vision‑enhanced math benchmarks while retaining strong performance on general‑domain evaluations (e.g., DocVQA, ChartQA, MMMU), indicating that math specialization need not sacrifice breadth.

\paragraph{Future work.}
We plan to (i) extend the pipeline to richer diagrammatic inputs and real‑world figures, (ii) explore adaptive and multi‑scale rendering policies conditioned on expression density. Integrating synthesized graphical data and broader instruction‑following diagnostics are additional priorities for improving robustness and usability.

\section*{Limitations}

Despite MathVLM's strong performance on both math-based and general-domain benchmarks, we note the following limitation:

\paragraph{Text-only rendered data.}
Our current pipeline focuses on improving multimodal reasoning using preprocessed text-only problems rendered as images. As a result, the training set underrepresents synthesized graphical content. We are working to integrate such graphical data into future training datasets.

\bibliography{custom}

\appendix

\section{Qualitative Examples of Model's Generation}
\label{sec:appendix}

Examples of the URSA-8B and MathVLM on the DocVQA task. We observe that URSA often engages in reasoning even when it is not explicitly requested by the question.


\begin{table*}[t!]
  \centering
  \small
  \caption{DocVQA prompts with answers from \textsc{MathVLM} and \textsc{URSA}.}
  \label{tab:docvqa-mathvlm-ursa}
    \resizebox{\linewidth}{!}{%
    \begin{tabular}{@{}p{0.2\linewidth} p{0.3\linewidth} p{0.3\linewidth}@{}}
      \toprule
      \textbf{Prompt (DocVQA)} & \textbf{MATHVLM Answer} & \textbf{URSA Answer} \\
      \midrule
      What is the \emph{Mortality Rate} in the year 1975 according to the chart? Answer with a single word of phrase. &
      \texttt{0.28} &
      \begin{minipage}[t]{\linewidth}\ttfamily
      Step 1: Observe the graph provided, focusing on the line representing ``Mortality Rate'' and the year 1975.\\
      Step 2: Locate the point on the ``Mortality Rate'' line that corresponds to the year 1975 on the horizontal axis.\\
      Step 3: Trace a vertical line from this point to the vertical axis, which represents the ``Mortality Rate'' values.\\
      Step 4: Read the value where the vertical line intersects the vertical axis. The value is 0.28.\\
      Answer: 0.28
      \end{minipage}
      \\
      \bottomrule
    \end{tabular}%
  }
\end{table*}


\end{document}